# Groupwise Constrained Reconstruction for Subspace Clustering

**Ruijiang Li**[†]                                                                                                   RJLI@FUDAN.EDU.CN
**Bin Li**[‡]                                                                                                      BIN.LI-1@UTS.EDU.AU
**Ke Zhang**[†]                                                                                                 K_ZHANG@FUDAN.EDU.CN
**Cheng Jin**[†]                                                                                                        JC@FUDAN.EDU.CN
**Xiangyang Xue**[†]                                                                                              XYXUE@FUDAN.EDU.CN

[†] School of Computer Science, Fudan University, Shanghai, 200433, China
[‡] QCIS Centre, FEIT, University of Technology, Sydney, NSW 2007, Australia

## Abstract

Reconstruction based subspace clustering methods compute a self reconstruction matrix over the samples and use it for spectral clustering to obtain the final clustering result. Their success largely relies on the assumption that the underlying subspaces are independent, which, however, does not always hold in the applications with increasing number of subspaces. In this paper, we propose a novel reconstruction based subspace clustering model without making the subspace independence assumption. In our model, certain properties of the reconstruction matrix are explicitly characterized using the latent cluster indicators, and the affinity matrix used for spectral clustering can be directly built from the posterior of the latent cluster indicators instead of the reconstruction matrix. Experimental results on both synthetic and real-world datasets show that the proposed model can outperform the state-of-the-art methods.

## 1. Introduction

Subspace clustering aims to group the given samples into clusters according to the criterion that samples in the same cluster are drawn from the same linear subspace. In the last decade, a number of subspace clustering methods have been proposed with successful applications in the areas including motion segmentation (Kanatani, 2001; Vidal & Hartley, 2004; Elhamifar & Vidal, 2009), image clustering under different illuminations (Ho et al., 2003), etc. Generally speaking, existing approaches to subspace clustering can be classified into the following categories: matrix factorization based, algebraic based, statistically modelling, and reconstruction based, among which the reconstruction based approach has been proved most effective and has drawn much attention recently (Elhamifar & Vidal, 2009; Liu et al., 2010; Wang et al., 2011). In this paper, we focus on the reconstruction based approach.

The objective of the reconstruction based subspace clustering is to approximate the dataset $\boldsymbol{X} \in \mathbb{R}^{D \times N}$ ($N$ is the number of samples and $D$ denotes the sample dimensionality) with the reconstruction $\boldsymbol{XW}$, where $\boldsymbol{W} \in \mathbb{R}^{N \times N}$ is the reconstruction matrix which can be further used to build the affinity matrix $|\boldsymbol{W}| + |\boldsymbol{W}^\top|$ for spectral clustering. The intuition behind the reconstruction is to make the value of $w_{ij}$ small or even vanish if samples $\boldsymbol{x}_i$ and $\boldsymbol{x}_j$ are not in the same subspace, such that the subspaces/clusters can be easily identified by the subsequent spectral clustering.

All the existing reconstruction based methods come with proofs claiming that the desired $\boldsymbol{W}$ could be obtained under the *subspace independence assumption*, i.e., the underlying subspaces $\mathcal{S}_1, \mathcal{S}_2, \cdots, \mathcal{S}_K$ are linearly independent, or mathematically,

$$\dim(\sum_{k=1}^{K} \oplus \mathcal{S}_k) = \sum_{k=1}^{K} \dim(\mathcal{S}_k) \qquad (1)$$

Unfortunately, this assumption will be violated if there exist bases shared among the subspaces. For example, given three orthogonal bases, $\boldsymbol{b}_1, \boldsymbol{b}_2, \boldsymbol{b}_3$, and two subspaces, $\mathcal{S}_1 = \boldsymbol{b}_1 \oplus \boldsymbol{b}_2$ and $\mathcal{S}_2 = \boldsymbol{b}_3 \oplus \boldsymbol{b}_2$ ($\boldsymbol{b}_2$ is shared in $\mathcal{S}_1$ and $\mathcal{S}_2$), the l.h.s. of Eq.(1) is 3, which is smaller than the r.h.s. being 4. In real-world scenarios, the subspace independence assumption does not always hold. For example, in human face clustering, as the number of clusters (persons) increases, the r.h.s. of Eq.(1) will exceed the l.h.s., which is upper bounded by the dimensionality of "human faces", so the subspace independence assumption will be violated eventually. Figure 1 illustrates this phenomenon based on the Extended Yale Database B (Georghiades et al., 2001). Once the subspace independence assumption is violated, there is no guarantee that the existing reconstruction based methods are able to obtain the desired $W$. In practice, we observe that the subspace independence assumption is critical to the success of the existing reconstruction based methods. Once the subspace independence assumption is violated, the performance of these existing reconstruction based methods become far from decent, even though the dimensionality of the underlying subspaces is low (shown in Section 4.1.1).

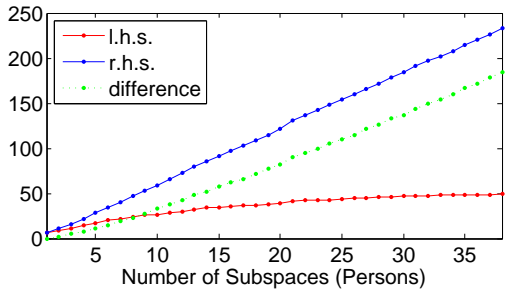

*Figure 1.* Evaluation of Eq.(1) on the Extended Yale Database B. The dimensionality is computed as the minimal number of principle components keeping 95% energy. Obviously, as the number of subspaces increases ($x$-axis), the r.h.s. of Eq.(1) grows linearly, while the l.h.s. is approaching a value upper bounded by the dimensionality of the combined space.

To tackle the subspace clustering problem, we propose a Groupwise Constrained Reconstruction (GCR) model, with the advantage that GCR no longer relies on the subspace independence assumption. In GCR, the sample cluster indicators are introduced as latent variables, conditioned on which the Slab-and-Spike-like priors are used as groupwise constraints to suppress the magnitude of certain entries in $W$. Thanks to these constraints, the requirement of the subspace independence assumption is no longer needed to obtain the desired $W$. Our method significantly differs from the existing methods in that, the reconstruction in GCR incorporates the information that "the samples can be grouped into clusters"; whereas in the existing methods, this information is ignored and the reconstruction depends solely on the data.

Another advantage of GCR is that, the affinity matrix needed for spectral clustering can be built from the cluster indicators rather than $W$. In our model, the reconstruction matrix $W$ can be analytically marginalized out. We first use Gibbs Sampler to collect samples from the posterior of the cluster indicators, then use the collected samples to build the "probabilistic affinity matrix", which is finally input to the spectral clustering algorithm to obtain the final clustering result. Compared with $|W| + |W^\top|$, which is used as the affinity matrix in the existing methods, the probabilistic affinity matrix built from the cluster indicators is more sophisticated, because it is naturally positive, symmetric and of clear interpretation. The experimental results on synthetic dataset, motion segmentation dataset and human face dataset show that GCR can outperform the state-of-the-art.

## 2. Background

In this section, we give a brief introduction to the previous works on subspace clustering.

### 2.1. Non-Reconstruction Based

Matrix factorization based methods Costeira & Kanade (1998); Kanatani (2001) approximate the data matrix with the product of two matrices, one containing the bases and the other containing the factors. The final clustering result is obtained by exploiting the factor matrix. These methods are not robust to noise and outliers and will fail if the subspaces are dependent.

The algebraic based General Principle Component Analysis (GPCA) (Vidal et al., 2005) fits the samples with a polynomial, with the gradient of a point orthogonal to the subspace containing it. This approach makes fewer assumptions on the subspaces, and the success is guaranteed when certain conditions are met. The major problem of the algebraic based approach is that the computational complexity is high (exponential to the number of subspaces and their dimensions), which restricts its application scenarios. In (Rao et al., 2010), Robust Algebraic Segmentation (RAS) is proposed to handle the data with outliers, but the complexity issue still remains.

Statistical models assume that the samples in each subspace are drawn from a certain distribution such as Gaussian, and take different objectives to find the optimal clustering result. For example, Mixture of Probabilistic PCA (Tipping & Bishop, 1999) uses the Expectation Maximization (EM) algorithm to find the maximum likelihood over all the samples, $k$-subspaces method (Ho et al., 2003) alternates between assigning the cluster to each sample and updating the subspaces, Random Sample Consensus (RANSAC) (Fischler & Bolles, 1981) keeps looking for the samples in the same subspace until the number of samples in the subspace is sufficient, then continues searching another subspace after removing these samples. Agglomerative Lossy Compression (ALC) (Ma et al., 2007) searches the latent subspaces by minimizing an objective containing certain information criteria with an agglomerative strategy.

## 2.2. Reconstruction Based

Reconstruction based methods usually consist of the following two steps: 1) Find a reconstruction for all the samples, in the form that each sample is approximated by the weighted sum of the other samples in the dataset. The optimization problem in Eq.(2) is solved to get the reconstruction weight matrix $\boldsymbol{W}$.

$$\min_{\boldsymbol{W}} \quad \ell(\boldsymbol{X} - \boldsymbol{X}\boldsymbol{W}) + \omega\Omega(\boldsymbol{W}) \qquad (2)$$
$$\text{s.t.} \quad w_{ii} = 0$$

where the term $l(\cdot) : \mathbb{R}^{D \times N} \mapsto \mathbb{R}$ measures the error made by approximating $\boldsymbol{x}_i$ with its reconstruction $\sum_{j \neq i} w_{ji} \boldsymbol{x}_j$, the term $\Omega(\cdot) : \mathbb{R}^{N \times N} \mapsto \mathbb{R}$ is used for regularization, and $\omega$ is a tradeoff parameter. 2) Apply spectral clustering algorithm to get the final clustering result from the reconstruction weights $\boldsymbol{W}$. Usually, $|\boldsymbol{W}| + |\boldsymbol{W}|^\top$ is treated as the affinity matrix input to the spectral clustering methods.

The methods of this class distinguish from each other in employing different regularization terms, i.e., $\Omega(\boldsymbol{W})$ in Eq.(2). In Sparse Subspace Clustering (SSC) (Elhamifar & Vidal, 2009), the authors propose to use the $l_1$ norm $\|\boldsymbol{W}\|_1$ to enforce the sparseness in $\boldsymbol{W}$, in the hope that the sparse coding process could shrink $w_{ji}$ to zero if $\boldsymbol{x}_i$ and $\boldsymbol{x}_j$ are not in the same subspace. In Low-Rank Representation (LRR) (Liu et al., 2010), nuclear norm $\|\boldsymbol{W}\|_*$ is used to encourage $\boldsymbol{W}$ to have a low rank structure[1], and $l_{2,1}$ norm is used as the $\ell(\cdot)$ term in Eq.(2) to make the method more robust to outliers. In SSQP (Wang et al., 2011), the authors choose $\Omega(\boldsymbol{W}) = \|\boldsymbol{W}^\top \boldsymbol{W}\|_1$, meanwhile force $\boldsymbol{W}$ to be non-negative. As a consequence, the optimization problem in Eq.(2) turns out to be a quadratic programming problem, for which the projected gradient descend method can be used to find a solution.

## 3. Groupwise Constrained Reconstruction Model

Consider a clustering task in which we want to group $N$ samples, denoted by $\boldsymbol{X} = [\boldsymbol{x}_1, \boldsymbol{x}_2, \cdots, \boldsymbol{x}_N] \in \mathbb{R}^{D \times N}$, into $K$ clusters, where $N$ is the number of samples, $D$ is the sample dimensionality, and $\boldsymbol{x}_i \in \mathbb{R}^D$ denotes the $i$-th sample. Let $\boldsymbol{z} = [z_1; z_2; \cdots; z_N]$ be the cluster indicator vector, where $z_i \in \{1, 2, \cdots, K\}$ indicates that sample $\boldsymbol{x}_i$ is drawn from the $z_i$-th cluster. The goal of subspace clustering is to find the cluster indicators $\boldsymbol{z}$, such that for each $k \in \{1, 2, \cdots, K\}$, the samples in the $k$-th cluster, i.e., $\{\boldsymbol{x}_i | z_i = k\}_{i=1}^N$, reside in the same linear space. This objective is quite different from the objective of traditional clustering methods, in which the variance of inter-cluster samples are minimized, such as K-means; or the "difference" of clusters are maximized, such as Discriminative Clustering (Ye et al., 2007).

---
[1] Nuclear norm $\|\boldsymbol{W}\|_*$ is defined to be the sum of singular values of matrix $\boldsymbol{W}$.

### 3.1. Model

Following the idea of the reconstruction based approach to subspace clustering, the Groupwise Constrained Reconstruction (GCR) model uses $p(\boldsymbol{X}|\boldsymbol{W})$ in Eq.(3) to quantify the reconstruction,

$$p(\boldsymbol{X}|\boldsymbol{W}, \boldsymbol{\sigma}) = \prod_{i=1}^{N} \mathcal{N}(\boldsymbol{x}_i | \sum_{j \neq i} w_{ji}\boldsymbol{x}_j, \sigma_i^2) \qquad (3)$$

where $\mathcal{N}(\cdot|\mu, \Sigma)$ denotes the Gaussian distribution with mean $\mu$ and variance $\Sigma$, $w_{ji}$ is the element at the $j$-th row, $i$-th column of matrix $\boldsymbol{W} \in \mathbb{R}^{N \times N}$, $\sigma_i^2 > 0$ is a random variable measuring the reconstruction error for the $i$-th sample, and $\boldsymbol{\sigma} = [\sigma_1; \sigma_2; \cdots; \sigma_N] \in \mathbb{R}^N$. We place an inverse Gamma prior on all the $\sigma_i$'s:

$$p(\sigma_i^2) = \text{IG}(\sigma_i^2 | \frac{\nu}{2}, \frac{\nu\lambda}{2}) \qquad (4)$$

where IG denotes the inverse Gamma distribution, and $\nu > 0$ and $\lambda > 0$ are given hyperparameters.

What makes the GCR model different is that, *GCR explicitly requires every sample to be reconstructed mainly by the samples in the same cluster*. In other words, the magnitudes of weights for the samples in different clusters should be small. Intuitively, $\boldsymbol{W}$ should be nearly block-wise diagonal if the samples are rearranged in a proper order (see Figure 2(b) for an illustration). To enforce such property of $\boldsymbol{W}$, we treat the cluster indicators $\boldsymbol{z}$ as latent random variables, and introduce a prior for $\boldsymbol{W}$ conditioned on $\boldsymbol{z}$ and $\boldsymbol{\sigma}$ as follows,

$$p(\boldsymbol{W}|\boldsymbol{z}, \boldsymbol{\sigma}) = \prod_{i=1}^{N} \prod_{j=1}^{N} \mathcal{N}(w_{ji}|0, \sigma_i^2 \alpha_{ji}) \qquad (5)$$

$$\alpha_{ji} = \alpha_{ij} = \begin{cases} \alpha_L & z_j \neq z_i \\ \alpha_H & z_j = z_i \end{cases}$$

where $\alpha_H > \alpha_L \geq 0$ are hyperparameters and $\frac{\alpha_L}{\alpha_H}$ is small. This prior is quite similar to the Slab and Spike prior used for variable selection (George & Mcculloch, 1997), with $\alpha_H$ corresponding to the slab and $\alpha_L$ corresponding to the spike. As the effects of Eq.(5), to generate $\boldsymbol{W}$ given the latent cluster indicators, if $\boldsymbol{x}_i$ and $\boldsymbol{x}_j$ are not in the same cluster/subspace, $w_{ji}$ and $w_{ij}$ are restricted to be small or close to the mean value 0 of the corresponding Gaussian distribution; if $\boldsymbol{x}_j$ and $\boldsymbol{x}_i$ come from the same cluster/subspace, the values of $w_{ji}$ and $w_{ij}$ could be either small or big. We make $\boldsymbol{W}$ dependent on $\boldsymbol{\sigma}$ as well, so that both $\boldsymbol{\sigma}$ and $\boldsymbol{W}$ can be further marginalized out by combining Eqs.(3), (4) and (5), which will be discussed later.

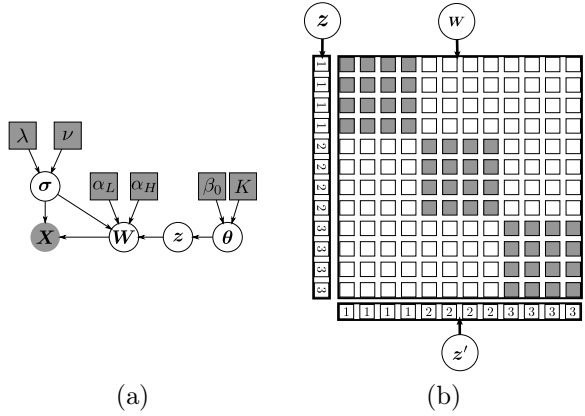

*Figure 2.* (a) Graphical representation for the GCR model. Squares and Circles denote parameters and random variables, respectively. Grey and White indicate observed (given) variables and latent variables, respectively. (b) Block-wise diagonal property of matrix $\boldsymbol{W}$ when samples are ordered such that samples from the same cluster/subspace are together. White cells denote the entries with small value associated with hyperparameter $\alpha_L$, and grey cells denote the entries with either small or big values associated with hyperpamameter $\alpha_H$.

Furthermore, we introduce a discrete prior

$$p(\boldsymbol{z}|\boldsymbol{\theta}) = \prod_{i=1}^{N} \text{Cate}(z_i|\boldsymbol{\theta})$$

for the cluster indicators $\boldsymbol{z}$ conditioned on $\boldsymbol{\theta} = [\theta_1; \theta_2; \cdots; \theta_K] \in \mathbb{R}^K$, where $\text{Cate}(z_i|\boldsymbol{\theta}) = \theta_{z_i}$ denotes the categorical distribution, and $\theta_k \in [0,1]$ can be viewed as the prior knowledge about the proportion of samples in the $k$-th cluster. Since it is difficult to set $\boldsymbol{\theta}$ beforehand, we use a Dirichlet distribution

$$p(\boldsymbol{\theta}) = \text{Dir}(\boldsymbol{\theta}|\frac{\beta_0}{K}\boldsymbol{1}_K)$$

as a prior for $\boldsymbol{\theta}$, where $\text{Dir}(\cdot)$ denotes the Dirichlet distribution, and $\boldsymbol{1}_K = [1, 1, \cdots, 1] \in \mathbb{R}^K$.

The hierarchical representation for GCR model is shown in Figure 2(a), and the full probability can be written as follows,

$$p(\boldsymbol{X}, \boldsymbol{W}, \boldsymbol{z}, \boldsymbol{\theta}, \boldsymbol{\sigma})$$
$$= [p(\boldsymbol{\theta})p(\boldsymbol{z}|\boldsymbol{\theta})] [p(\boldsymbol{W}|\boldsymbol{z}, \boldsymbol{\sigma})p(\boldsymbol{X}|\boldsymbol{W}, \boldsymbol{\sigma})p(\boldsymbol{\sigma})] \quad (6)$$

Observing that $\boldsymbol{W}, \boldsymbol{\sigma}$ and $\boldsymbol{\theta}$ in Eq.(6) can be marginalized out analytically, we can write down $p(\boldsymbol{z}|\boldsymbol{X})$, de- noted as $q(\boldsymbol{z})$ for short, as follows,

$$q(\boldsymbol{z}) \propto f_0 \prod_{i=1}^{N} f_i \quad (7)$$

$$f_0 = \prod_{k=1}^{K} \Gamma\left(\frac{\beta_0}{K} + n_k(\boldsymbol{z})\right)$$

$$f_i = \det(\boldsymbol{C}_i)^{-\frac{1}{2}} \left(\boldsymbol{x}_i^\top \boldsymbol{C}_i^{-1} \boldsymbol{x}_i + \nu\lambda\right)^{-\frac{D+\nu}{2}}$$

$$\boldsymbol{C}_i = \boldsymbol{H}_{z_i} - \alpha_H \boldsymbol{x}_i \boldsymbol{x}_i^\top$$

$$\boldsymbol{H}_k = \sum_{j|z_j=k} \alpha_H \boldsymbol{x}_j \boldsymbol{x}_j^\top + \sum_{j|z_j\neq k} \alpha_L \boldsymbol{x}_j \boldsymbol{x}_j^\top + \boldsymbol{I}_D$$

where $f_0$ and $f_i$ comes from the first and the second brackets in Eq.(6), respectively; $n_k(\boldsymbol{z})$ is the number of samples in the $k$-th cluster; $\Gamma(\cdot)$ denotes the Gamma function; and $\boldsymbol{I}_D \in \mathbb{R}^{D\times D}$ denotes the identity matrix.

### 3.2. Obtaining the Final Clustering Result

We use the Gibbs Sampling algorithm (MacKay, 2003) to approximate the posterior distribution $q(\boldsymbol{z})$. In each epoch, for $i \in \{1, 2, \cdots, N\}$, the Gibbs sampler iteratively updates $z_i$ to a sample drawn from $p(z_i|z_{\sim i}, \boldsymbol{X}) = \frac{p(\boldsymbol{z}|\boldsymbol{X})}{p(z_{\sim i}|\boldsymbol{X})} \propto q(\boldsymbol{z})$, where $z_{\sim i} = \{z_j | j \neq i\}$. A direct implementation will lead to the time complexity of $\mathcal{O}(N^2 D^3)$ for each epoch. Fortunately, the complexity can be reduced to $\mathcal{O}(N^2 D + KD^2)$ using rank-1 update. At the end of each epoch, we collect the values of all the cluster indicators as a sample of $\boldsymbol{z}$. Finally, we save the samples of $\boldsymbol{z}$ from the last $M$ epochs, denoted as $\boldsymbol{s}_1, \boldsymbol{s}_2, \cdots, \boldsymbol{s}_M$, and discard the samples left. We can use the following two approaches to obtain the final clustering result.

**MAP approach.** Use the last collected sample $\boldsymbol{s}_M$ as an initialization, then maximize the posterior $q(\boldsymbol{z})$ in Eq.(7) by alternating among $z_1, z_2, \cdots, z_N$. The local maximum is directly used as the clustering result.

**Bayesian approach.** With the collected samples, we first compute an affinity matrix $\boldsymbol{G}_m \in \mathbb{R}^{N\times N}$ over the $N$ samples, where

$$(\boldsymbol{G}_m)_{ij} = \begin{cases} 1 & (\boldsymbol{s}_m)_i = (\boldsymbol{s}_m)_j \\ 0 & (\boldsymbol{s}_m)_i \neq (\boldsymbol{s}_m)_j \end{cases} \quad (8)$$

then compute the "probabilistic affinity matrix" $\boldsymbol{G} = \frac{1}{M}\sum_m \boldsymbol{G}_m$; and finally put $\boldsymbol{G}$ into a classical clustering method to obtain the final clustering result.

Here, $G_{ij}$ can be treated as an approximation to the posterior distribution $p(z_i = z_j|\boldsymbol{X})$. Compared with existing reconstruction based methods which use $|\boldsymbol{W}| + |\boldsymbol{W}|^\top$ as the affinity matrix input into the spec-tral clustering algorithm, our probabilistic affinity matrix $G$ is more sophisticated since $G_{ij}$ can be clearly interpreted as the the possibility that sample $i$ and $j$ share the same cluster label. What is more, our affinity matrix $G$ is naturally positive and symmetric, whereas $|W| + |W|^\top$ is somehow like an ad-hoc way to "force" $W$ to be an affinity matrix.

### 3.3. When $K \to +\infty$

From Eq.(8) we see that, to obtain the probabilistic affinity matrix $G$, it is not mandatory to set $K$ to be the exact number of subspaces. In fact, the probabilistic affinity matrix can be obtained with any positive integer $K$. Particularly, we are interested in the GCR model when $K$ goes to positive infinity, in which case, the number of non-empty clusters remains a finite number (at most $N$ when each sample forms its own cluster). This strategy is in analogy with the Infinite Gaussian Mixture Model with Dirichlet Process (Rasmussen, 1999). For this reason, we refer to the GCR model with $K \to +\infty$ as GCR-DP. As the limit of GCR, the posterior of $z$ for GCR-DP is

$$\hat{q}(z) \propto \hat{f}_0 \prod_{i=1}^{N} f_i, \ \hat{f}_0 = \beta_0^{\hat{K}-1} \prod_{k=1}^{\hat{K}} \Gamma(n_k(z)) \quad (9)$$

where $f_i$ remains the same as in Eq.(7), and $\hat{K}$ is the number of non-empty clusters [2]. The Gibbs sampling procedure is similar to that of the original GCR, and the difference is described as follows. Suppose for now there are $K'$ non-empty clusters, to update $z_i$, besides computing $K'$ values for the non-empty clusters by plugging $z_i \leftarrow \{1, 2, \cdots, \hat{K}\}$ into the r.h.s. of Eq.(9), we need to compute an extra value for a new empty cluster by plugging $\hat{K} \leftarrow K' + 1$ and $z_i = K' + 1$ into Eq.(9). Then the Categorical sampler picks a cluster indicator for $z_i$ according to these $K' + 1$ values. If the indicator for the new cluster $(K'+1)$ is picked, we create a new empty cluster and put the $i$-th sample into it. The variables for the empty clusters can be removed to save the computational resource.

In the case of $K \to \infty$, Eq.(6) shows that there exists a trade-off among the reconstruction quality, prior for the cluster indicators and $p(W|z, \sigma)$. $p(W|z, \sigma)$ prefers more clusters, in which case more spikes in Eq.(5) could be introduced into the model, resulting in high p.d.f. of $p(W|z, \sigma)$. On the contrary, the Dirichlet process prior favors fewer number of clusters. In the premise of good reconstruction quality ($p(X|W, \sigma)$ is high), the competition between the Dirichlet process prior $p(z)$ and $p(W|z, \sigma)$ provides a way to circumvent the trivial solutions to the model (all the samples in one cluster or each sample in it's own cluster).

Due to the allowance to create more clusters, the outliers, which cannot be well reconstructed by the inliers, have the chance to "stand alone". As a result, the influence of the outliers can be reduced.

### 3.4. Hyperparameters

$\beta_0$: Throughout our experiment, $\beta_0$ for the Dirichlet distribution is always set to 1.

$\lambda$ and $\nu$: From Eq.(4) we see that $\lambda$ and $\nu$ control the reconstruction quality. According to the property of the inverse Gamma distribution, we have $\mathbb{E}(\sigma_i^{-1}) = \frac{1}{\lambda}$ and $\text{Var}(\sigma_i^{-1}) = \frac{2}{\lambda^2 \nu}$. Thus, it is reasonable to set $\lambda$ to a smaller number if the dataset are less noisy, and set $\nu$ to a smaller number if the variance of the reconstruction quality for different samples is higher (e.g., the dataset has more outliers). In our experiments, these two parameters are tuned for different datasets.

$\alpha_H$ and $\alpha_L$: According to Eq.(5), $\sigma_i^2 \alpha_H$ and $\sigma_i^2 \alpha_L$ directly influence the magnitude of $w_{ji}$. Since $\mathbb{E}(\sigma_i^{-1}) = \frac{1}{\lambda}$, we can use $\lambda \alpha_H$ and $\lambda \alpha_L$ to control the magnitude of $w_{ji}$ intuitively. After integrating out $\sigma$, we can rewrite the prior for $W$ as $p(W|z) = \prod_{i=1}^{N} \prod_{j=1}^{N} \mathcal{T}(w_{ji}|\nu, 0, \lambda \alpha_{ji})$, where $\mathcal{T}(\cdot|u, v, w)$ denotes the student $t$ distribution with degree of freedom $u$, mean $v$ and variance $w$. Therefore, it is natural to use the mean value of the $t$ distribution to control the magnitude of $W$. In practice, we find that $\lambda \alpha_H = 0.1$ and $\frac{\alpha_H}{\alpha_L} = 10000$ yield good performance.

## 4. Experimental Results

In this section, we compare our methods with the other three reconstruction based subspace clustering methods: LRR (Liu et al., 2010), SSC (Elhamifar & Vidal, 2009) and SSQP (Wang et al., 2011). In our evaluation, the quality of clustering is measured by accuracy, which is computed as the maximum percentage of match between the clustering result and the ground truth. For GCR, the MAP estimation is directly used as the final clustering result; for GCR-DP, we first compute the probabilistic affinity matrix according to Eq.(8), then use NCut (Shi & Malik, 2000) to get the final clustering result. For MCMC, we treat $G^{(0)} = \left|\left(X^\top X + \delta I\right)^{-1}\right| \in \mathbb{R}^{N \times N}$ as the affinity matrix, and the result of spectral clustering is used as the initialization[3]. This can be understood by switching the rule between sample ($N$) and dimension ($D$), in such a way that $\boldsymbol{G}^0$ becomes the precision matrix over $N$ samples, and $\left|\boldsymbol{G}_{ij}^{(0)}\right|$ measures the dependency between the $i$-th and the $j$-th samples conditioned on the other samples. We set the number of epochs for the Gibbs sampler to 500, and use the last 100 samples to construct the probabilistic affinity matrix. We find that under such settings, our methods runs faster than SSC and SSQP empirically.

---

[2] To use Eq.(9), $z$ should be reorganized so that the first $\hat{K}$ clusters are non-empty.

[3] $\delta$ is a jitter value making the matrix invertible.

### 4.1. Synthetic Datasets

We use synthetic datasets to investigate how these reconstruction based methods perform when the subspace independence assumption mentioned in Section 1 is violated. The synthetic data containing $K$ subspaces are generated as follows: 1) Generate a matrix $\boldsymbol{B} \in \mathbb{R}^{2 \times 50}$, each column of which is drawn from a Gaussian distribution $\mathcal{N}(\cdot|0, \boldsymbol{I}_2)$. 2) For the $k$-th cluster containing $n_k$ samples, generate $\boldsymbol{y}_1 \in \mathbb{R}^{n_k}$, the elements of which are drawn independently from the uniform distribution defined on $[-1, 1]$. After that, generate $\boldsymbol{y}_2 = \tan\frac{16k}{17K}\boldsymbol{y}_1$ (avoiding $\tan\frac{\pi}{2}$). Finally, generate the $n_k$ samples in the $k$-th cluster as $[\boldsymbol{y}_1, \boldsymbol{y}_2]\boldsymbol{B} \in \mathbb{R}^{n_K \times 50}$. All the experiments here are repeated for 5 times.

#### 4.1.1. VIOLATION OF SUBSPACE INDEPENDENCE ASSUMPTION

For $K = 2, 3, \cdots, 8$, we generate 7 datasets according to the steps listed above. For these synthetic datasets, the l.h.s. of Eq.(1) is 2, and the r.h.s. of Eq.(1) is $K$. Thus, the degree of the violation of the subspace independence assumption increases as $K$ increases. The results are reported in Figure 3(a).

As we can see, LRR and SSC perform well when the subspace independence assumption holds ($K = 2$) or is slightly violated ($K = 3$). However, their performance decreases significantly as the violation degree increases, even though their parameters are tuned for different $K$. In contrast, GCR and GCR-DP are able to retain high performance even though the violation degree keeps increasing.

In the case of $K = 8$, we compare the affinity matrices produced by these reconstruction based methods, as shown in Figure 4. Obviously, the affinity matrix produced by GCR-DP has stronger discrimination power on the clusters than those of the others. The affinity matrix produced by SSQP looks promising. However, a deep investigation shows that in the matrix the sum of many rows are zero, making the clustering performance less satisfactory.

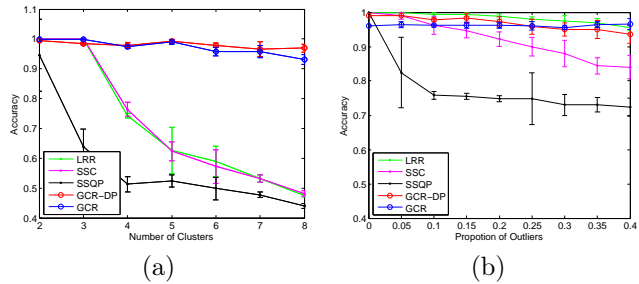

*Figure 3.* Comparison on the Synthetic Datasets. (a) Accuracy on the synthetic dataset when subspace independence assumption is violated. (b) Accuracy on the synthetic dataset with increasing portion of noisy samples.

#### 4.1.2. INCREASING PORTION OF NOISY SAMPLES

Consider the case when there exist samples deviating from the exact positions in the subspaces. Following the previous listed steps, we generate a dataset containing 2 subspaces, each of which contains 50 samples. We add Gaussian noises $\mathcal{N}(\cdot|0, 3)$ to $0\%, 5\%, \cdots, 40\%$ of the samples, respectively. The results on the 9 datasets are reported in Figure 3(b).

The results show that our methods and LRR are able to maintain high accuracy even though high portion of the samples deviate from their ideal position. The success of LRR is due to the $l_{2,1}$ norm used for the loss term in Eq.(2), while the success of GCR and GCR-DP may be due to the model in which each sample has its own parameter $\sigma_i$ to measure the reconstruction error. SSC performs less better, and its performance remains acceptable when the noise level is low.

### 4.2. Hopkins 155 Dataset

We evaluate our models on the Hopkins 155 motion dataset. This dataset consists of 155 sequences, each of which contains the coordinates of about $39 - 550$ points tracked from 2 or 3 motions. The task is to group the points into clusters according to their motions for each sequence. Since the coordinates of the points from a single motion lie in an affine subspace with the dimensionality at most 4 (Elhamifar & Vidal, 2009), we project the coordinates in each sequence into $4r$ dimensions with PCA, where $r$ is the number of motions in the sequence, then append 1 as the last dimension of each sample. The results are reported in Table 1. This dataset contains a small number of latent subspaces, and the results of the compared methods have no significant difference.

### 4.3. MSRC Dataset

In the MSRC dataset, 591 images are provided with manually labeled image segmentation results (each re-

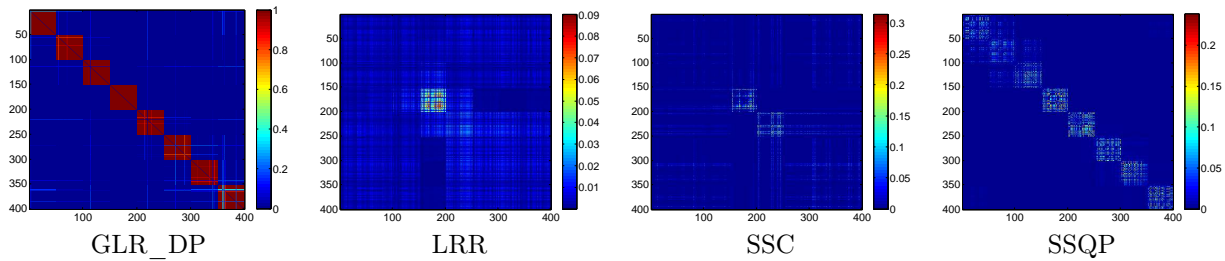

Figure 4. The affinity matrices produced by reconstruction based methods. The samples are ordered so that samples from the same cluster are adjacent.

Table 1. Accuracy on the Hopkins 155 Dataset

| Method | Mean | Median | Min |
|---|---|---|---|
| LRR | .9504 | .9948 | .5820 |
| SSC | .9729 | 1 | .5766 |
| SSQP | .9536 | 1 | .5450 |
| GCR-DP | .9764 | 1 | .5532 |
| GCR | .9608 | .9970 | .5833 |

Table 2. Accuracy on the MSRC Dataset with 459 Images

| Method | Mean | Median | Min |
|---|---|---|---|
| LRR | .6625 | .6500 | .3514 |
| SSC | .6548 | .6400 | .3673 |
| SSQP | .6550 | .6374 | .3784 |
| GCR-DP | .6651 | .6667 | .3587 |
| GCR | .7046 | .6964 | .3838 |

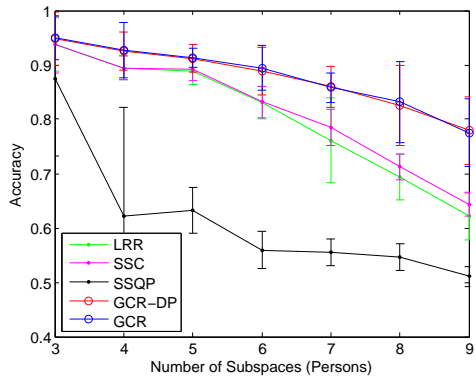

Figure 5. Results On Extended YaleB Dataset with Increasing Number of Subspaces.

### 4.4. Human Face Dataset

We also evaluate our method on the Extended Yale Database B (Georghiades et al., 2001). This database contains 2414 cropped frontal human face images from 38 subjects under different illuminations, and grouping these images can be treated as a subspace clustering problem, because it is shown in (Ho et al., 2003) that the images for a fixed face under different illuminations can be approximately modeled with low dimensional subspace. To evaluate the performance of all these methods, we form 7 tasks, each of which contains the images from randomly picked $\{3, 4, \cdots, 9\}$ subjects, respectively. We resize the images to $42 \times 48$, then use PCA to reduce the dimensionality of the raw features to 30. We repeat the experiment for 5 times and show the results in Figure 5.

gion is given a label, and there are totally 23 labels). Following (Cheng et al., 2011), for each image, we group the superpixels, which are small patches in an *over-segmented* result, with subspace clustering methods. The groundtruth (cluster label) for a superpixel is given as the label of region it belongs to.

In our experiment, 100 superpixels are extracted for each image with the method described in (Mori et al., 2004), and each superpixel is represented with the RGB Color Histogram feature of dimensionality 768. We discard all the superpixels with label "background", and then discard the images containing only one label. Finally, we get 459 images. For each image, the average number of superpixels is 91.3, and the number of clusters ranges from 2 to 6. We use PCA to reduce the dimensionality to 20 in order to keep 95% energy. The results are show in Table 2.

Clearly, our methods outperform the other three on this dataset. GCR also performs better than GCR-DP because it utilizes the information about the number of latent subspaces during the reconstruction step.

The performance of GCR and GCR-DP are better than the other three methods. In particular, with the number of subspaces increasing, the difference between the l.h.s. and r.h.s. of Eq.(1) increases (see Figure 1). Consequently, the performance of LRR, SSC and SSQP, which rely on the subspace independence assumption to build the affinity matrix, degrades quickly. On the contrary, GCR and GCR-DP utilize the information that "the samples can be grouped into subspaces", thus they are less influenced by the violation of subspace independence assumption.

## 5. Conclusion and Discussion

We propose the Groupwise Constrained Reconstruction (GCR) models for subspace clustering in this paper. Compared with other reconstruction based methods, our models no longer rely on the subspace independence assumption, which usually gets violated in the applications in which the number of subspaces keeps increasing. On the synthetic datasets, we show that existing reconstruction based methods suffer from the violation of the subspace independence assumption, while the affinity matrix produced by our model, which is built from the posterior of the latent cluster indicators, is more sophisticated and of stronger discrimination power on discovering the latent clusters. On the three real-world datasets, our methods show promising results.

Besides the subspace clustering problem, the idea of groupwise constraints can be further applied to other problems involving graph construction. For example, in semi-supervised learning (SSL), the constraints can be modified such that a sample is only allowed to be reconstructed by its neighbors in the Euclidean space. In this way, the cluster assumption and manifold assumption, which are two fundamental SSL assumptions, can be neatly unified within our framework. For dimension reduction methods such as LLE (Roweis & Saul, 2000), it is also interesting to design new models to use the posterior of the reconstruction matrix for embedding, such that the local and global structure of the data could be preserved simultaneously.

### Acknowledgements

We thank the anonymous reviewers for valuable comments. This work was partially supported by 973 Program (2010CB327906), Shanghai Leading Academic Discipline Project (B114), Doctoral Fund of Ministry of Education of China (20100071120033), and Shanghai Municipal R&D Foundation (08dz1500109). Bin Li thanks UTS Early Career Researcher Grants.